\documentclass[sigconf]{acmart}
\renewcommand\footnotetextcopyrightpermission[1]{} 
\usepackage{booktabs} 
\usepackage{algorithm}
\usepackage{algorithmic}
\usepackage{subcaption}
\usepackage{graphicx}
\usepackage{pgfplots}
\usepackage{float}

\pgfplotsset{compat=1.14}

\setcopyright{rightsretained}

\acmDOI{placeholder}

\acmISBN{placeholder}

\acmConference{The Web Conference 2018}{April 2018}{Lyon, France} 
\acmYear{2018}
\copyrightyear{2018}

\acmArticle{4}
\acmPrice{15.00}

\begin{document}
\title{Surfacing contextual hate speech words within social media}

\author{Jherez Taylor}
\authornote{These authors contributed equally to the paper}
\affiliation{%
    \institution{National Tsinghua University}
    \city{Hsinchu} 
    \country{Taiwan}
}
\email{jherez.taylor@gmail.com}

\author{Melvyn Peignon}
\authornotemark[1]
\affiliation{%
    \institution{National Tsinghua University}
    \city{Hsinchu} 
    \country{Taiwan}
}
\email{melvyn.peignon@gmail.com}

\author{Yi-Shin Chen}
\affiliation{%
    \institution{National Tsinghua University}
    \city{Hsinchu} 
    \country{Taiwan}
}
\email{yishin@gmail.com}


\begin{abstract}
Social media platforms have recently seen an increase in the occurrence of hate speech 
discourse which has led to calls for improved detection methods. Most of these
rely on annotated data, keywords, and a classification technique. 
While this approach provides good coverage, it can fall short when dealing 
with new terms produced by online extremist communities which act as original 
sources of words which have alternate hate speech meanings. These code words
(which can be both created and adopted words) are designed
to evade automatic detection and often have benign meanings in regular discourse. 
As an example, \textit{``skypes, googles, yahoos''} are all instances of words 
which have an alternate meaning that can be used for hate speech. This overlap 
introduces additional challenges when relying on keywords for both the collection of data
that is specific to hate speech, and downstream classification. In this work, we 
develop a community detection approach for finding extremist hate speech 
communities and collecting data from their members. We also develop a word 
embedding model that learns the alternate hate speech meaning of words and 
demonstrate the candidacy of our code words with several annotation
experiments, designed to determine if it is possible to recognize a word as being
used for hate speech without knowing its alternate meaning. We report an inter-annotator
agreement rate of $K=0.871$, and $K=0.676$ for data drawn from our
extremist community and the keyword approach respectively, 
supporting our claim that hate speech detection is a contextual task and does not
depend on a fixed list of keywords. Our goal is to advance the domain by providing 
a high quality hate speech dataset in addition to learned code words that can 
be fed into existing classification approaches, thus improving the accuracy of 
automated detection.
\end{abstract}

%
%

\ccsdesc[500]{Computing methodologies~Natural language processing}
\ccsdesc[300]{Computing methodologies~Unsupervised learning}
\ccsdesc{Human-centered computing~Social media}

\keywords{hate speech, community detection, NLP, social media}

\maketitle

\newcommand*{\HW}{$\mathcal{H} $ }
\newcommand*{\HComm}{HateComm }
\newcommand*{\TClean}{TwitterClean }
\newcommand*{\THate}{TwitterHate }
\newcommand*{\simByWord}{$simByWord$ }
\newcommand*{\WM}{$\mathcal{W} $ }
\newcommand*{\DM}{$\mathcal{D} $ }
\newcommand*{\EM}{$\mathcal{E} $ }
\newcommand*{\EMvc}{$\mathcal{E}_{vc} $ }
\newcommand*{\WMc}{$\mathcal{W}_{C} $}
\newcommand*{\WMh}{$\mathcal{W}_{H} $}
\newcommand*{\DMc}{$\mathcal{D}_{C} $}
\newcommand*{\DMh}{$\mathcal{D}_{H} $}
\newcommand*{\CG}{$\mathcal{CG} $ }
\newcommand*{\GS}{$\mathcal{CG}_{similarity} $ }
\newcommand*{\GR}{$\mathcal{CG}_{relatedness} $ }
\newcommand*{\topn}{$topn$}

\section{Introduction}

The internet allows for the free flow of information and one of its major growing
pains has been the propagation of hate speech and other abusive content, however,
it is becoming increasingly common to find hateful messages that attack a person or a
group because of their nationality, race, religion or gender. Sentences like
\textit{I fucking hate niggers} or \textit{go back to your muslim shithole}
\footnote{\textbf{Reader advisory}: We present several examples that feature hate speech and
explicit content. We want to warn the reader that these examples are lifted from
our data set and are featured here for illustrative purposes only.} can
be readily found even when viewing topics that should be far removed from hate speech.
This creates an atmosphere that becomes uncomfortable to engage in and can have a significant
impact on online discourse and it inflicts a damaging financial
and social cost on both the social network and the victims alike. Twitter has reportedly lost business 
partially as a result of potential buyers raising concerns about the reputation that
the social network has for bullying and uncivil communication.~\cite{BLOOM}.
Additionally, the European Union has moved to enact a law that will impose hefty
fines on social media networks that fail to remove flagged hate speech content 
within 24 hours, and other offensive content within 7 days, even going as far as
to hold personal staff accountable for the inaction of these companies. ~\cite{forbes}

To address the issue, social networks like Twitter try to balance the need to
promote free speech and to create a welcoming environment. The Terms of Service
for these platforms provide guidelines on what content is prohibited,
these guidelines then shape the automatic filtering tools of these platforms. 
However, Hate Speech [HS] can be difficult to define as there are some who argue that 
restrictions on what constitutes HS are in fact violations of the right to free speech. 
The definition can also vary in terms of geographic location and the laws that can be applied.
It is thus important to adhere to a rigid definition of HS in our work. 

For this work we rely on the definition from the \emph{International Covenant
on Civil and Political Rights, Article 20 (2)} which defines Hate Speech as
\emph{any advocacy of national, racial or religious hatred that constitutes
incitement to discrimination, hostility or violence}~\cite{ICCPR}. 
In a troubling development, online communities of users that engage in HS 
discourse are constantly crafting new linguistic means of bypassing automatic filters.
These include intentional misspellings and adapting common words to have alternative meanings,
effectively softening their speech to avoid being reported and subsequently 
banned from the platform. There are two major challenges that need to be considered:
\begin{itemize}
  \item \textbf{Substitution}: members of online hate speech communities tend to
  to substitute words that have accepted hate speech meanings with something that
  appears benign and out of context, to be only understood by fellow community 
  members. This is not unlike the use of codewords for open communication. 
  To illustrate, consider the following example, 
  \textit{``Anyone who isn't white or christian does not deserve to live in the
  US. Those foreign \underline{skypes} should be deported."} Here, the word ``skypes''
  is a code word used to refer to Jewish people. The example would likely be 
  missed by a classifier trained with word collocation features, as it does
  not contain any words strongly associated with hate speech, a problem
  highlighted by Nobata et al.\cite{Nobata2016}. We can infer that ``skypes''
  is being used as a code word here and we can also infer possible words that are
  both similar and substitutable such as ``niggers'' or ``muslims''.\\
  
  \item \textbf{Non representative data}: Keyword sampling is often used to collect
  data but those keywords often overlap with many topics. For example, there is
  no distinction between the words \textit{fuck, fucking, shit}, which are often
  used for hate speech as well as regular conversations. Extensive
  annotation is first required before any methodology can be applied. Additionally,
  Some users also limit what they say in public spaces and instead link to 
  extremist websites that express their shared ideas, minimizing
  their risk being banned. This creates a certain
  fuzziness that has so far not been fully addressed when using public data for 
  hate speech research.
\end{itemize}

In this paper we aim to develop a method that detects hate speech communities 
while also identifying the hate speech codewords that are used to avoid detection.
We make use of word dependencies in order to detect the contexts in which words
are utilized so as to identify new hate speech \textit{code words} that might not exist in the
known hate speech lexicon. Specifically, to address the challenges outlined, 
this paper has the following contributions:

\begin{itemize}
  \item We develop a graph based methodology to collect hateful content shared by extremist communities.
  \item We address the constant introduction of new hate speech terms with our 
  contextual word enrichment model that learns out-of-dictionary hate speech \textit{code words}
  \item We make public our dataset and our code word pipeline as a means to expand
  existing hate speech lexicon.
\end{itemize}

Our results show the benefit of collecting data from hate speech communities for
use in downstream applications. We also demonstrate the utility of considering syntactic
dependency-based word embeddings for finding words that function
similar to known hate speech words (code words). We present our work as  an
online system that continuously learns these dependency embeddings, thus
expanding the hate speech lexicon and allowing for the retrieval of more tweets
where these code words appear.

\section{Related Work}

The last several years has seen an increase in research related to identifying
HS within online platforms, with respect to both hate speech classification
and the detection of extremist communities. O'Callaghan et al.~\cite{ocallahan}
made use of Twitter profiles to identify and analyse the relationships between members of extremist 
communities which consider cross-country interactions as well. They note that
linguistic and geographic proximity influences the way in which extremist 
communities interact with each other. Also central to the problem that we attempt to 
solve is the idea of supplementing the traditional bag of words [BOW] approach. 
Burnap and Williams~\cite{Burnap2016}
introduced the idea of \textit{othering} language (the idea of differentiating groups of with
``us" versus ``them" rhetoric) as a useful feature for HS
classification. Long observed in discussions surrounding racism and HS, their work
lends credence to the idea that HS discourse is not limited to the presence or 
absence of a fixed set of words, but instead relies one the context in which it appears.
The idea of out-of-dictionary HS words is a key issue in all related classification
tasks and this work provides us with the basis and motivation for constructing a
dynamic method for identifying these words. However, hate speech detection is a difficult
task as it is subjective and often varies between individuals. Waseem~\cite{Waseem2016B}
speaks to the impact that annotators have on the underlying classification models.
Their results show the difference in model quality when using expert versus 
amateur annotators, reporting an inter-annotator agreement of $K=0.57$ amateurs
and $K=0.34$ for the expert annotators. The low scores indicated that hate speech
annotation and by extension classification, is difficult task and represents a 
significant and persistent challenge.

Djuric et al.~\cite{Djuric2015} adopted the \emph{paragraph2vec} [a modification
of \textit{word2vec}] approach for classifying user comments as being either 
abusive or clean. This work was extended by Nobata et
al.~\cite{Nobata2016}, which made use of features from \textit{n}-grams, Linguistic, 
Syntactic and Distributional Semantics. These features form their model, \emph{comment2vec},
where each comment is mapped to a unique vector in a matrix of representative words.
The joint probabilities from word vectors were then used to predict the next word in a
comment. As our work focuses on learning the different contexts in which words
appear, we utilize neural embedding approaches with \textit{fasttext} by
Bojanowski et al.~\cite{DBLP:journals/corr/BojanowskiGJM16} and
\textit{dependency2vec} by Levy and Goldberg~\cite{omer2014}.

Finally, Magu et al.~\cite{Magu2017} present their work on detecting hate speech
code words which focused on the manual selection of hate speech code
words. These represent words that are used by extremist communities to spread hate content
without being explicit, in an effort to evade detection systems. A fixed seed of 
code words was used to collect and annotate tweets where those
words appear for classification. These code words have an accepted
meaning in the regular English language which users exploit
in order to confuse others who may not understand their hidden meaning. In contrast
to this work, we propose our method for dynamically
identifying new code words. All of the previous studies referenced here utilize either an initial bag of 
words [BOW] and/or annotated data and the general consensus is that a BOW alone is not sufficient. 
Furthermore, if the BOW remains static then trained models would struggle to classify less explicit HS
examples, in short, we need a dynamic BOW. 

To advance the work, we propose the use of hate speech community detection in 
order to get data which fully represents how these communities use words for 
hate speech. We use this data to obtain the different types of textual context 
as our core features for surfacing new hate speech code words. This context 
covers both the topical and functional context of the words being used. 
The aim of our work is to dynamically identify new \textit{code words} 
that are introduced into the corpus and to minimize the reliance on a BOW and 
annotated data.
\section{Background}

\subsection{Addressing Hate Speech Challenges}

Firstly, we must define our assumptions about hate speech and the role that
context takes in our approach. Our goal is to obtain data from online 
hate speech communities, data which can be used to build models that create word
representations of \textit{relatedness} and \textit{similarity}. 
We present our rationale for collecting data from online hate speech communities
and explain the various types of context used throughout our methodology.

While there exists words or phrases that are known to be associated with hate
speech\footnote{We used lists scraped from
http://www.hatebase.org/} as used by Nobata et al.~\cite{Nobata2016}.,
it can often be expressed without any of these keywords. Additionally,
it is difficult for human annotators to identify hate speech if they are not
familiar with the meaning of words or any context that may surround the text as
outlined in \cite{Waseem2016A}. These issues make it difficult to identify hate 
speech with Natural Language Processing [NLP] approaches. Further compounding the
issue of hate speech detection, the members of these online communities have 
adopted strategies for bypassing the automatic detection
systems that social networks employ. One such strategy being used is word substitution,
where explicit hate speech words are replaced with benign words which have hidden
meanings. Ultimately, the issue with code words is one of word polysemy and it is
particularly difficult to address because these alternate meanings do not exist
in the public lexicon.

To deal with the problem of code word substitution, we use word \textit{similarity}
and word \textit{relatedness} features to train contextual representations of words that
our model can use to identify possible hate speech usage. To do this, it is 
necessary to use models that align words into vector space in
order to get the neighbours of a word under different uses. These models are
referred to as Neural Embeddings and while most in the same fundamental way, the
distinction comes from the input (hereafter referred to as \textit{context})
that they make use of. We introduce \textit{topical context} and
\textit{functional context} as key concepts that will influence our features.

\subsection{Neural Embeddings and Context}

Neural Embeddings refer to the various NLP techniques used for mapping words or phrases to
dense vector representations that allow for efficient computation of semantic
similarities of words. The idea is based on Distributional Hypothesis by
Harris~\cite{Harris1981} which states that \textit{``words that appear in the
same contexts share semantic meaning"}, meaning that a word shares characteristics
other words that are typically its neighbours in a sentence. Cosine similarity 
is the measure used for vector similarity, it will hereafter appear as $sim$.
Neural Embedding models represent words in
vector space. Given a target word $w$, an embedding model $\mathcal{E}$, 
it and a specified $topn$ value, it is possible to retrieve the
$topn$ most $sim$ words in $\mathcal{E}$, \simByWord will be used to reference 
this function hereafter.

Topical Context is the context used by word embedding approaches like \textit{word2vec}~\cite{NIPS2013_5021},
that utilize a bag-of-words in an effort to rank words by their domain similarity.
Context here is considered as the window for each word in a sentence, the
task being to extract target words and their surrounding words (given a window size)
to predict each context from its target word. In doing so it models word \textit{relatedness}.
However, functional context describes and ranks words by the syntactic relationships
that a word participates in. Levy and Goldberg \cite{omer2014} proposed a method of adapting
\textit{word2vec} to capture the Syntactic Dependencies in a sentence with
\textit{dependency2vec}. Intuitively, Syntactic Dependencies refers to the word
relationships in a sentence. Such a model might tell us words close to \textit{Florida},
words might be \textit{New York, Texas, California}; words that reflect that 
\textit{Florida} is a state in the United States. We simplify this with the 
term \textit{similarity}, to indicate that words that share similar 
functional contexts are similar to each other.

Functional context is modelled by \textit{dependency2vec}, a modification of 
\textit{word2vec} proposed by Levy and Goldberg~\cite{omer2014} who build the
intuition behind Syntactic Dependency Context. The goal of the
model is to create learned vector representations which reveal words that are
functionally similar and behave like each other, i.e., the model captures word
\textit{similarity}. \textit{dependency2vec} operates in the same way as
\textit{word2vec} with the only difference being the representation of
\textit{context}. The advantage of this approach is that the model is then able
to capture word relations that are outside of a linear window and can thus
reduce the weighting of words that appear often in the same window as a target
word but might not actually be related.

\textit{Topical context} reflects words that associate with each other
(\textit{relatedness}) while \textit{functional context} reflects words that
behave like each other (\textit{similarity}). In our work we wish answer the
following: \textit{how do we capture the meaning of code words that we do not
know the functional context of?} To provide an intuitive understanding and motivation for the use of both
\textit{topical} and \textit{functional} context we provide an example. Consider
the following real document drawn from our data:

\begin{tabbing}
$v_{nt}$ \= = \= value \kill
\textit{Skypes and googles must be expelled from our homelands} \\
\end{tabbing}

\begin{table}[ht]
\centering
\caption{Comparing word context results}
\label{table:relsim}
\begin{tabular}{lll}
                           & \multicolumn{2}{c}{\textbf{skypes}}                                                                                                                                                     \\
\textbf{Clean Texts} & \multicolumn{1}{l|}{\begin{tabular}[c]{@{}l@{}}skyped\\ facetime\\ skype-ing\\ phone\end{tabular}} & \begin{tabular}[c]{@{}l@{}}whatsapp\\ line\\ snapchat\\ imessage\end{tabular}      \\ \cline{2-3}
\textbf{Hate Texts}   & \multicolumn{1}{l|}{\begin{tabular}[c]{@{}l@{}}chat\\ dropbox\\ kike\\ line\end{tabular}}          & \begin{tabular}[c]{@{}l@{}}cockroaches\\ negroes\\ facebook\\ animals\end{tabular} \\
                           & \textbf{Relatedness}                                                                               & \textbf{Similarity}
\end{tabular}
\end{table}

With the example we generate Table \ref{table:relsim} which displays the top 4 
words closest to the target word \textit{skypes}, across two different datasets
and word contexts. We assume the existence of embedding
models trained on relatively clean text and another trained on text filled with
hate speech references. For words under the \textit{relatedness} columns,
we see that they refer to internet companies. In this case, while we know
that \textit{skypes} is a hate speech code word it still appears alongside the 
internet company words because of the word substitution problem.
We see the same effect for \textit{similarity} under Clean Texts. 
However, when looking at \textit{similarity} under the Hate Texts we can 
infer that the author is not using Skype in its usual form. The \textit{similarity} 
columns gives us words that are functionally similar. 
We do not yet know what the results mean but anecdotally we see that the model 
returns groups of people and it is this type of result we wish to exploit in 
order to detect code words within our datasets. It is for this reason that we 
desire Neural Embedding models that can learn both word \textit{similarity} 
and word \textit{relatedness}. We propose that this can be used as an 
additional measure to identify unknown hate speech \textit{code words} that are 
used in similar \textit{functional contexts} to words that already have defined
hate speech meanings.
\section{Methodology}

\subsection{Overview}
The entire process consists of four main steps:

\begin{itemize}
    \item Identifying online hate speech communities
    \item Creating Neural Embedding models that capture word relatedness and word similarity.
    \item Using graph expansion and PageRank scores to bootstrap our initial HS seed words.
    \item Enriching our bootstrapped words to learn out-of-dictionary terms that
    bare some hate speech relation and behave like code words.
\end{itemize}

The approach will demonstrate the effectiveness of our hate speech community detection process.
Additionally, we will leverage existing research that confirmed the utility of using
hate speech blacklists, syntactic features, and various neural embedding
approaches. We provide a an overview of our community detection methodology, 
as well as the different types of word context, and how they can be utilised to 
identify possible code words.

\subsection{Extremist Community Detection}

A key part of our method concerns the data and the way in which it was collected
and partitioned, as such it is important to first outline our method and
rationale. There exists words that can take on a vastly different
meanings depending on the way in which they are used, that is, they act as codewords
under different circumstances. Collecting data from extremist communities which
produce hate speech content is necessary to build this representation. There are
communities of users on Twitter and elsewhere that share a high proportion of 
hate speech content amongst themselves and it is reasonable to expect that they 
would want to share writing or other content that they
produced with like minded individuals. We are of the belief that new hate speech codewords are created by
these communities and that if there was any place to build a dataset that reflects a 
``hate speech community'' it would be at the source. We began the search by 
referencing the Extremist Files maintained by the Southern Poverty Law Center[SPLC]\footnote{https://www.splcenter.org/}, 
a US non-profit legal advocacy organization that focuses on civil rights issues and litigation.

The SPLC keeps track of prominent extremist groups and individuals within the
US, including several websites that are known to produce
extremist and hate content, most prominent of these being
DailyStormer\footnote{https://www.dailystormer.com/} and American Renaissance\footnote{https://www.amren.com/}. 
The articles on these websites are of a White Supremacist nature and are filled 
with references that degrade and threaten non white groups, as such, 
it serves as an ideal starting point for our hate speech data collection. The two
websites mentioned were selected as our seed and we crawled their articles,
storing the author name, the article body, and its title. The list of 
authors was then used for a manual lookup in order to tie the article author to 
their Twitter account. We were not able to identify the profile of each author as
some of the accounts in our list self identified as being pseudo-names. 
For each of these Twitter accounts we extracted their followers and friends, 
building an oriented graph where each vertex represents a user and edges represent
a directional user-follower relationship. In order to discover authors that were
missed during the initial pass, we use the centrality betweenness of different 
vertices to get prominent users. Due to preprocessing constraints we opted to compute 
an approximate betweenness centrality.
\begin{definition}(Vertices)\label{VerticesGraphDef}
For this relationship graph, $V$ refers to the set containing all vertices while $V'$ is a random
subset of $V$. We utilize SSSP (single source, shortest path) which is defined as $s,t \in V'$, 
the number of shortest paths from $s$ to $t$, $\sigma_{st}$. Similarly, the number
of shortest paths between $s$ and $t$ going through $v$, $\sigma_{st}(v)$ is thus:
\begin{displaymath}
    \forall v \in V', g(v) = \sum\limits_{s \neq t}\sum\limits_{t \neq s}{\frac{\sigma_{st}(v)}{\sigma_{st}}}
\end{displaymath}
\end{definition}

The computed betweenness centrality for every element in $V'$ is then used to 
extrapolate the value of other nodes, as described in Brandes et.al~\cite{brandes2007}.
From there, an extended seed of a specified size will be selected based on the 
approximated centrality of the nodes and the original author. With this extended seed,
it becomes possible to collect any user-follower relationships that were initially missed.
After the initial graph processing, over 3 million unique users IDs were obtained.
A random subset of vertices was then taken to reduce the size of the graph for 
computational considerations. This random subset forms a graph $\mathcal{G}$ 
containing $V_f$ vertices, $\vert V_f \vert\approx 20000$. Each vertex of $\mathcal{G}$ represents
a user while directed edges represent relationships. Consider
$s, t \in V$, if $s$ is following $t$ then a directed edge $(s,t)$ will exists.
Historical tweet data was collected from these vertices, representing over 36 millions tweets.
We hereafter refer to graph $\mathcal{G}$ as \textbf{HateComm, our dataset which consists
of the article content and titles previously mentioned in addition to the historical 
tweets of users within the network of author followers.}

The issue with code words is that they are by definition secret or at best, not
well known. Continuing with the examples of \textit{Skype} and \textit{Google}
we previously introduced, if we were to attempt to get related or even similar
words from a Neural Embedding model trained on generalized data, it is unlikely
that we would observe any other words that share some relation to hate speech.
However, it is not enough to train models on data that is dense with hate speech.
The results might highlight a relation to hate speech but would provide no information
on the frequency of use in different situations, in short, we need to have some 
measure of the use of a word in the general English vocabulary
in order to support the claim that these words can also act as hate speech code
words. It is for this reason that we propose a model that includes word
\textit{similarity, relatedness} and frequency of use, drawn from the differing
datasets. We therefore introduce two additional datasets that we collect from
Twitter, the first using hate speech keywords and the second collected from the
Twitter stream without any search terms. Twitter offers a free 1\% sample of 
the total tweets sent on the platform and so we consider tweets collected in 
this manner to be a best effort representation of the average.

\begin{figure}[H]
\includegraphics[clip,width=0.90\columnwidth,height=5.0cm]{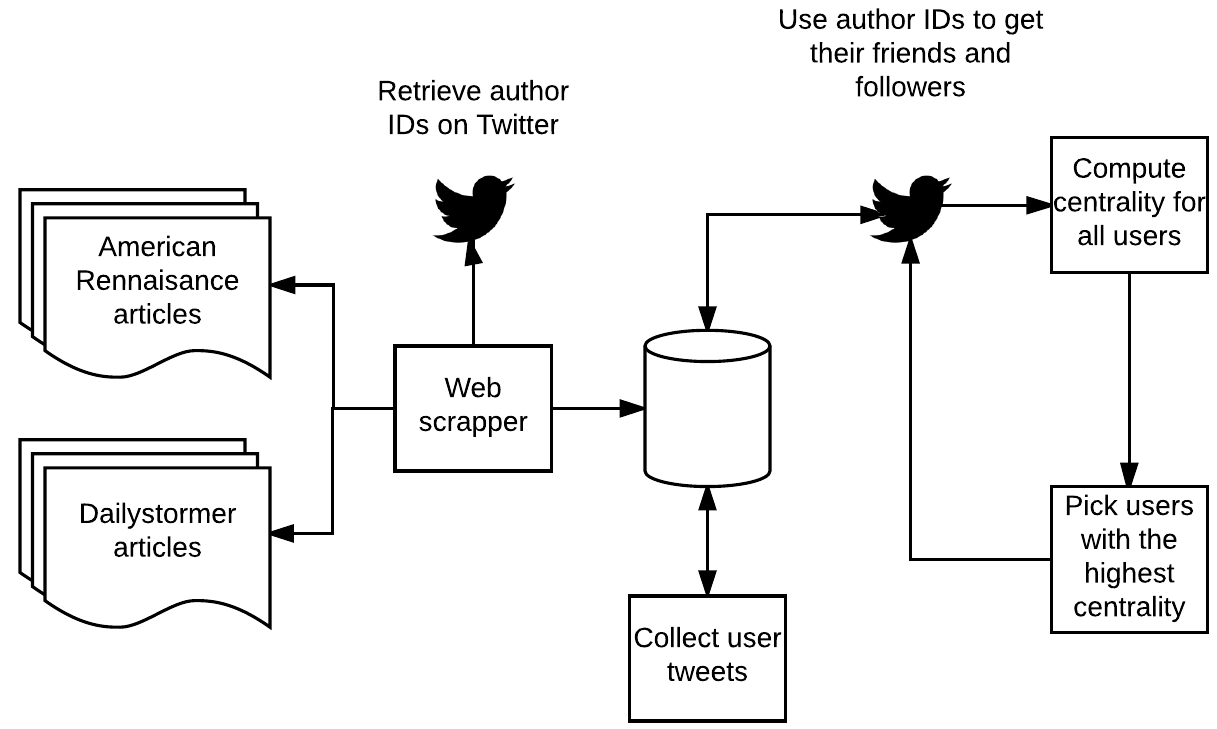}
\caption{Data flow to collect the tweets}
\label{community_flow}
\end{figure}

\textit{Hate Speech Keywords} is is defined as a set of words 
\HW = $\left\{\text{$h_1,..,h_n$}\right\}$ typically associated with hate speech in the English 
language. We made use of the same word source as~\cite{Nobata2016}.
\textbf{TwitterHate refers to our dataset of tweets collected using \HW as seed words.}
While \textbf{TwitterClean refers to our dataset collected without tracking any specific terms or users},
only collecting what Twitter returned, free from the bias of collecting data
based on keywords. We filter and remove any tweet that contains a word $w \in$
$\mathcal{H}$.

\subsection{Contextual Code Word Search}

For our work we dynamically generate contextual word representations which 
we use for determining if a word acts as a hate speech code word or not.
To create contextual word representations we use the Neural Embedding models 
proposed by \textit{dependency2vec}\cite{omer2014} and
\textit{fasttext}~\cite{DBLP:journals/corr/BojanowskiGJM16}.
As we wish to identify out-of-dictionary words that can
be linked to hate speech under a given context, as part of our preprocessing we 
we then define a graph based approach to reduce the word search space. Finally,
our method for highlighting candidate code words is presented. We report our code words
as well as the strength of the relationship that they may have to hate speech.

\subsubsection{Embedding Creation}
Creating a model that align words into vector space allows for the
extraction of the neighbours of a word under different uses. Our intuition is 
that we can model the \textit{topical} and \textit{functional}
context of words in our hate speech dataset in order to identify
out-of-dictionary hate speech code words. For our \HComm and \THate datasets we
create both a Word Embedding Model and a Dependency Embedding Model as
We refer to these as \WMc, \WMh, \DMc, and \DMh.

\begin{figure}[H]
    \includegraphics[trim={2.03cm 14cm 2.2cm 1.8cm},clip,width=1.32\columnwidth,height=9.7cm]{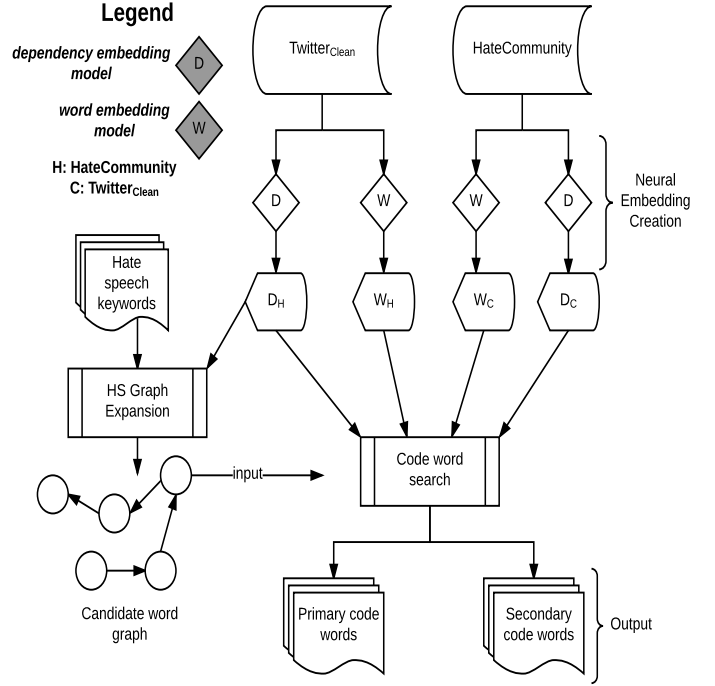}
    \caption{Framework Overview}
    \label{top_level_framework}
\end{figure}

\subsubsection{Contextual Graph Filtering}

The idea for finding candidate code words is based on an approach that considers
the output from the $topn$ word list from our 4 embedding models, given a target
word \textit{w}. Filtering the list of possible out of dictionary words is required
to reduce the search space and obtain non hate speech words input our code word search. 
to check. To achieve this, we devised a graph construction methodology that builds a 
weighted directed graph of words with the output from an embedding model. In this way,
we can construct a graph that models word \textit{similarity} or word \textit{relatedness},
depending on the embedding model we utilize. This graph takes on several different inputs and parameters
throughout the algorithm, as such we define the general construction.

\begin{figure}[H]
    \includegraphics[width=1.06\columnwidth]{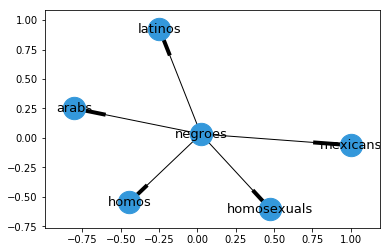}
    \caption{Graph $\mathcal{CG}_1$, built from word$_1$}
    \label{fig:graph1}
\end{figure}

\begin{definition} (Contextual Graph)\label{ContextualGraphDef} is a weighted
directed graph \CG where each vertex $v \in V$ represents a word $w \in
seed\_input$. Edges are represented by the set $E$. The graph represents word
\textit{similarity} or word \textit{relatedness}, depending on the embedding
model used at construction time. For a pair of vertices $(v_1, v_2)$ an edge $e \in E$ is created if $v_2$
appears in the output of \simByWord, with $v_1$ as the input word. As an intuitive
example, using $v_1$ = \textit{negroes} from Table \ref{table:relsim} the output 
contextual graph can be seen in Figure \ref{fig:graph1}.
\end{definition}

To further reduce the search space we use PageRank~\cite{ilprints422} to rank 
out-of-dictionary words in a graph where some of the vertices are known hate 
speech keywords. This allows us to model known hate speech words and words close 
to them as \textit{important links} that pass on their weight to their their successor
vertices, thus boosting their importance score. In this way we are able to have
the edges that are successors of a known hate speech word get a boost which 
reflects a higher relevance in the overall graph.

\begin{definition} (boost)\label{HSboostDef} During the construction of any
    \textit{contextual graph} we do a pre-initialization step where we call
    \simByWord with a given $topn$ for $\forall w \in$ \HW if $w \in$ \EMvc. Recall
    that \EMvc is the stored vocabulary for the embedding model used during graph
    construction. The frequency of each word in the resulting collection is stored
    in $boost$. $boost(w)$ thus returns the frequency of the word $w$ in this
    initialization step, if it exists.
\end{definition}
    
This boosting gives us words that are close to known hate speech keywords and
is done to to assign a higher weighting based on the frequency of a word in a list
generated with the \HW seed. Using cosine similarity scores alone as the edge weight
would not allow us to model the idea that hate speech words are the important ``pages" in 
the graph, the key concept behind PageRank. Concisely, this boosting is
done to set known hate speech words as the important ``pages" that pass 
on their weight during the PageRank computation. Edge attachment is then done via 2
weighting schemes that we employ.

\begin{definition}(weightingScheme)\label{weightDef}
    Let $frq(v)$ denote the frequency of vertex $v$ in \EMvc for the given embedding
    model, and $sim(v_1, v_2)$ the cosine similarity score for the embedding vectors
    under vertices $v_1$ and $v_2$. The weight $wt$ of e($v_1,v_2$) is then defined
    in the following:
    
    \begin{displaymath}
    \label{eqn:weight}
    	wt(v_1,v_2) = \begin{cases}
        \log(frq(v_1))\times boost(v_1) + sim(v_1, v_2) & \text{if } v_1 \in boost \\
        sim(v_1, v_2) & \text{if }  v_1 \notin boost
        \end{cases}
	\end{displaymath}
\end{definition}

With the prerequisite definitions in place we now outline our algorithm for
building an individual word contextual graph in Algorithm \ref{alg:wordgraph}.
Intuitively, the algorithm accepts a target word and attaches edges to vertices
that appear in the results for $simByWord$. We then collect all vertices in the
graph and repeat the process, keeping track of the vertices that we have seen.
Note that $depth$ specifies the number of times that we collect current graph
vertices and repeat the process of appending successor vertices. A $depth$ of
$2$ indicates that we will only repeat the process for unseen vertices twice.

\begin{algorithm}[H]
    \caption{buildGraph}
    \label{alg:wordgraph}
    \begin{algorithmic}[1]
        \renewcommand{\algorithmicrequire}{\textbf{Input:}}
        \renewcommand{\algorithmicensure}{\textbf{Output:}}
        \REQUIRE $w, \mathcal{E}, depth, boost, topn$
            \ENSURE $\mathcal{CG}$
        \STATE seen\_vertices = $\emptyset$
        \STATE $\mathcal{CG}$ = empty directed graph
        \STATE predecessor\_vertices = $simByWord(w, topn)$
        \FOR {vertex:$p$ in predecessor\_vertices}
            \STATE $\mathcal{CG}$ += $add\_edge(w, p, wt(w,p))$
        \ENDFOR
        \STATE seen\_vertices += $w$
        \FOR {$i$ in $range(1, depth)$}
            \STATE curr\_vertices $\gets \mathcal{CG}.vertices()$
            \FOR {vertex:$v$ in curr\_vertices}
                \IF{$v \notin seen\_vertices$}
                    \STATE successor\_vertices = $simByWord(v, \mathcal{E}, topn)$
                        \FOR {each vertex $p$ in predecessor\_vertices}
                            \STATE $\mathcal{CG}$ += $add\_edge(v, p, wt(v,p))$
                        \ENDFOR
                        \STATE seen\_vertices += $v$
                \ENDIF
                \ENDFOR
        \ENDFOR
        \STATE \textbf{return} $\mathcal{CG}$
    \end{algorithmic}
\end{algorithm}

Our hate speech seed graph $\mathcal{CG}$ then becomes a union of contextual graphs [\ref{ContextualGraphDef}]
created from a list of words, with a graph being created for each word.
We opted to use \textit{similarity} embedding model over \textit{relatedness} for this step. The
union can be seen in the following equation.

\begin{displaymath}
    \label{eqn:union}
    \mathcal{CG} = \bigcup_{w \in \mathcal{H}} buildGraph(w, \mathcal{D}, depth, boost, topn)
\end{displaymath}

We then perform PageRank on the hate speech seed graph and use the document 
frequency $df$ [$df = \frac{doc\_count(w)}{N}$] for a given word $w$ as a cut-off measure,
where $N$ is the total number of documents in a given dataset; subsequently removing 
all known hate speech words from the output. The assumption
is that if a word $w$ in our \HW Graph is frequently used as a
code word, then it should representing have a higher $df$ in \HComm over
\TClean. To illustrate, we wouldn't expect hate communities
to use the word \textit{animals} for it's actual meaning more than the 
general dataset. This assumption is supported by plotting the frequencies and observing
that most of the words in the \HW graph have a high $df$ in \HComm and
it is necessary to surface low frequency words. We perform
several frequency plots and the results confirm our assumption. For the PageRank scores we
set $d=0.85$ as it is the standard rate of decay used for the algorithm. $PR =
PageRank(\mathcal{CG}, d=0.85)$ and trim $PR$ as outlined in the equation:

\begin{displaymath}
    \label{eqn:prcutoff}
    	\begin{cases}
    		keep(w) & \text{if } df(w\in HateComm) > df(w \in TwitterClean) \\
    		remove(w) & \text{if } df(w \in HateComm) < df(w \in TwitterClean)
    	\end{cases}
\end{displaymath}

Finally, we further refine our seed list, by building a new graph using the
trimmed $PR + \mathcal{H}$, computing a revised $PR$ on the resulting graph. To
be clear, only the word in this list and not the actual scores are used as input
for our codeword search.

\begin{table}[H]
\centering
\caption{Notations}
\label{table:notations}
\begin{tabular}{@{}ll@{}}
Notation & Description                                                         \\ \midrule
\CG      & a contextual graph built with output from \EM                       \\
\DMc     & a learned $dep2vec$ model trained on \TClean                        \\
\DMh     & a learned $dep2vec$ model trained on \HComm                        \\
\EM      & a learned embedding model of type \WM or \DM                        \\
\EMvc    & a stored vocabulary for a given embedding model                     \\
\WMc     & a learned word embedding model trained on \TClean                   \\
\WMh     & a learned word embedding model trained on \HComm
\end{tabular}
\end{table}

\subsubsection{Contextual Code Word Search}\label{codewordsearch}

With our trimmed PageRank list as input, we outline our process for selecting
out-of-dictionary hate speech code words. We place words into categories which
represent words that may be very tightly linked to known hate speech words and 
those that have a weaker relation.

\begin{definition} (getContextRep)\label{getContextDef} At the core of the
method is the mixed contextual representation that we generate for an input word
$w$ from our \HComm and \TClean datasets. It simply gives us word the relatedness
and word similarity output from embedding models trained on \HComm.
The process is as follows:

\begin{displaymath}
    \label{eqn:contextrep1}
    cRep(w)_{HateSimilar} = simByWord(w, \mathcal{D}_{H}, topn)
    \end{displaymath}
    \begin{displaymath}
    \label{eqn:contextrep2}
    cRep(w)_{HateRelated} = simByWord(w, \mathcal{W}_{H}, topn)
    \end{displaymath}
\end{definition}

\begin{definition} (primaryCheck)\label{primaryCheckDef}
accepts a word $w$, its contextual representation, and $topn$ to determine if
$w$ should be placed in the primary code word bucket, returning true or false.
Here, primary buckets refers to words that have some strong relation to known
hate speech words. First we calculate thresholds which check whether the number of 
number of known hate speech words in the contextual representation for a 
given word is above the specified threshold $th$.

\begin{gather*}
    \label{eqn:thsim}
    th\_similarity = \left ( th \geqslant \frac{size(HW \bigcap cRep_{HateSimilar})}{topn} \right )\\
    th\_relatedness = \left ( th \geqslant \frac{size(HW \bigcap cRep_{HateRelated})}{topn} \right )
\end{gather*}

With both thresholds, we perform an OR operation with $th\_check = th\_similarity \lor th\_relatedness$.
Next, we determine whether $w$ has a higher frequency in \HComm over \TClean by $freq\_check = \left (df(w \in HateComm) > df(w \in TwitterClean) \right )$. Finally, a word is selected as a primary code word with $ primary = th\_check \land freq\_check$
\end{definition}
    
\begin{definition} (secondaryCheck)\label{secondaryCheckDef}
    accepts a word $w$ and its contextual graph \CG and searches the vertices for
    any $v \in$ \HW, returning the predecessor vertices of $v$ as a set if a match
    is found as well as. We check that the set is not empty and use the truth value
    to indicate whether $w$ should be placed in the secondary code word bucket.
    $secondary = predecessor\_vertices(v \in \mathcal{G} \Rightarrow v \in \mathcal{H})$
\end{definition}

\section{Experiment Results}

\subsection{Training Data}

In order to partition our data and train our Neural Embeddings we first
collected data from Twitter. Both \TClean and \THate are composites of data
collected over several time frames, including the two week window leading up to
the 2016 US Presidential Elections, the 2017 US Presidential Inauguration, and
at other points during early 2017, consisting of around 10M tweets each. In
order to create \HComm we crawled the websites obtained from the SPLC as
mentioned in Section 4.2 and obtained a list of authors and attempted to link
them to their Twitter profiles. This process yielded 18 unique profiles from
which we collected their followers and built a graph of user:followers. We then
randomly selected 20,000 vertices and collected their historical tweets,
yielding around 400K tweets. \HComm thus consists of tweets and the article
contents that were collected during the scraping stage.

We normalize user mentions as \emph{user\_mention}, preserve \textit{hashtags}
and \textit{emoji}, and lowercase text. The tokenizer built for Twitter in the
Tweet NLP\footnote{http://www.cs.cmu.edu/~ark/TweetNLP/} package was used. It
should be noted that the Neural Embeddings required a separate preprocessing
stage, for that we used the NLP package Spacy\footnote{https://spacy.io/} to
extract syntactic dependency features.

\subsection{Experimental Setup}

As mentioned previously, we utilized \textit{fasttext} and
\textit{dependency2vec} to train our Neural Embeddings. For our Dependency
Embeddings we used 200 dimension vectors and for fasttext we utilized 300.

To initialize our list of seed words for our approach we built a contextual
graph with the following settings.

\begin{enumerate}
\item \DMh was used to build a \CG based on \HW \textit{similarity}
\item to generate \textit{boost} we set $topn=20$
\item We consider singular and plural variations of each $w \in $ \HW
\item Vertices were added with $topn=3$ and $depth=2$
\end{enumerate}

This process for expanding our \HW seed returned $994$ words after trimming
with the frequency rationale. For the contextual code word search we used the following:
\begin{enumerate}
\item $depth=2$, $topn=5$, $th=0.2$
\end{enumerate}

We set $th=0.2$ after experiments showed that most words did not return more
than 1 known hate speech keyword when checking its 5 closest words. This process return 55
primary and 262 secondary bucket words. It should be noted that we filtered for
known \HW including any singular or plural variations. An initial manual
examination of this list gave the impression that while the words were not
directly linked to hate speech, the intent could be inferred under certain
circumstances. It was not enough the do a manual evaluation as we needed a way
to verify if the words we had surfaced could be recognized as being linked to
hate speech under the right context. We saw fit to design an experiment to test
our results.

\subsection{Baseline Evaluation}

\begin{figure}[H]
  \includegraphics[width=1.06\columnwidth]{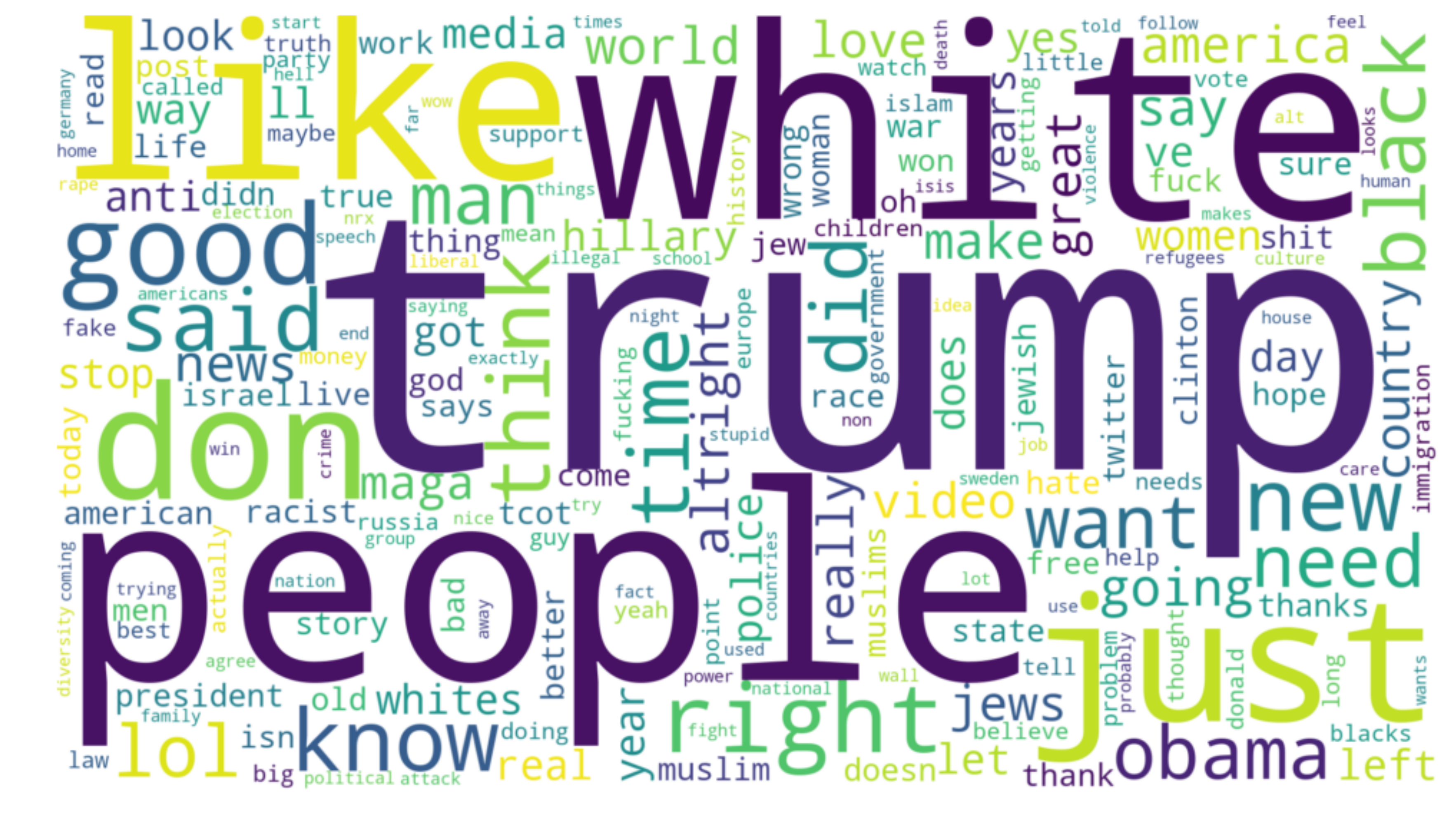}
  \caption{TF-IDF output}
  \label{fig:tfidf}
\end{figure}

\begin{figure}[H]
  \includegraphics[width=1.06\columnwidth]{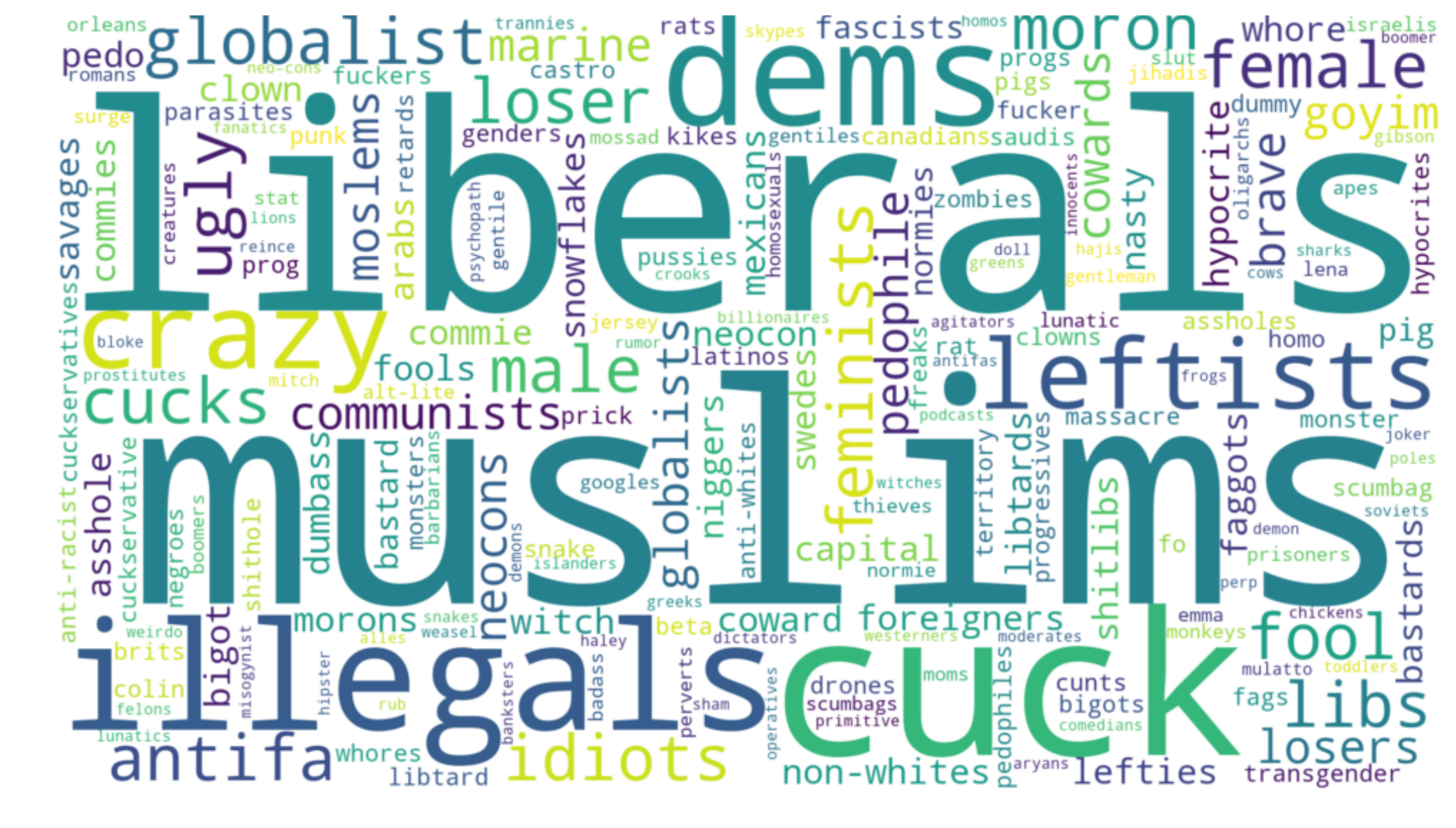}
  \caption{Contextual code word output}
  \label{fig:ctx}
\end{figure}

The major difficulty of our work has been choosing a method to evaluate our 
results as their are few direct analogues. As our baseline benchmark we calculate 
the tf-idf word scores for \HComm and compare with frequencies for our surfaced code words.
Using the tf-idf scores is a common approach for discovering the ideas present in
a corpus. Where higher tf-idf scores indicate a higher weight, due to
low frequency use of our code words lower scores represent a higher weight. For the
code word weights we use inverse document frequency. Figures \ref{fig:tfidf} and
\ref{fig:ctx} show the difference between the TF-IDF baseline and our contextual
code word search. The TF-IDF output appears to be of a topical nature, particularly
politics while the code word output features multiple derogatory references 
throughout.

The contextual code word approach is not without drawbacks, as ultimately these
words are a suggestion of possible hate speech code words. However, it represents
an improvement over attempting to find these hate speech code words manually as
the model can learn new hate speech code words as they are introduced.



\subsection{Annotation Experiment}

We have claimed throughout or work that context is important and we designed an
experiment to reflect that. Our aim was to determine if a selection of
annotators would be able to identify when a given word was being used in a hate
speech context without the presence of known hate speech keyword and without known the meaning
of the code words. The experiment featured manually selected code words
including 1 positive and 1 negative control word. It is important to have some
measure of control as many different works including ~\cite{Waseem2016B} have
highlighted the difficulty of annotating hate speech. The positive and negative
samples were designed to test if annotators could identify documents that
featured explicit hate speech (positive) and documents that were benign
(negative).

We built three distinct experiments where:

\begin{enumerate}
\item Documents refer to tweets and article titles.
\item 10 code words were manually selected and participants were asked to rate a document on a scale of very unlikely
(no references to hate speech) to very likely (hate speech) [0 to 4].
\item HateCommunity, TwitterClean, and TwitterHate were utilized as the sample pool, randomly
drawing 5 documents for each code word (10 word X 5 documents for each experiment).
\item Control documents were the same across all three experiments and did not feature known HS words apart from the positive control.
\item Direct links were only provided for the experiments drawn from HateCommunity and TwitterClean. After completing these experiments, participants were given the option
to move on to the \THate experiment.
\end{enumerate}

\begin{figure}[H]
\centering
\resizebox{.95\linewidth}{!}{
\begin{tikzpicture}
 \begin{axis}[title  = Aggregate percentage,
    xbar,
    y axis line style = { opacity = 0 },
    axis x line       = none,
    tickwidth         = 0pt,
    enlarge y limits  = 0.2,
    enlarge x limits  = 0.02,
    symbolic y coords = {Very unlikely, Unlikely, Neutral, Likely, Very likely},
    nodes near coords,
  ]
  \addplot coordinates { (0.743,Very unlikely) (0.1126,Unlikely)
                         (0.0969,Neutral)  (0.03,Likely) (0.016,Very likely)};
   \addplot coordinates { (0.07,Very unlikely) (0.07562,Unlikely)
                           (0.07464,Neutral)  (0.21137,Likely) (0.56833,Very likely)};
  \legend{water, niggers}
  \end{axis}
\end{tikzpicture}}
\caption{Aggregate percentage for control words}
\label{ctrl_words}
\end{figure}
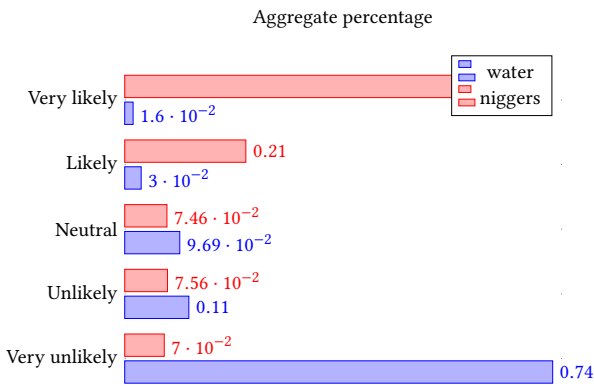

The experiment was designed to draw for our distinct datasets which would
reflect the use of the same word across differing situations and contexts. We
obtained 52, 53, and 45 responses for  HateCommunity, TwitterClean, and TwitterHate respectively. The
full list can be seen in Table \ref{table:experiment}. Table \ref{table:quesitons} provides a view of a few of the documents annotators
were asked to rate. None of the examples features known \HW but it is possible
to infer the intent of the original author. The experiment also featured control
questions designed to test if participants understood the experiment, we
provided 5 samples that featured the use of the word \textit{nigger} as positive
for hate speech and \textit{water} as negative for hate speech. An overwhelming
majority of the were able to correctly rate both control questions, as can be
seen in Figs. \ref{ctrl_words}.

\begin{table}[H]
\centering
\caption{Experiment Selection}
\label{table:experiment}
\begin{tabular}{@{}ll@{}}
\toprule
\multicolumn{2}{c}{code words}                                \\ \midrule
niggers {[}positive control{]} & water {[}negative control{]} \\
snake                          & googles                      \\
cuckservatives                 & skypes                       \\
creatures                      & moslems                      \\
cockroaches                    & primitives                  
\end{tabular}
\end{table}

\begin{table}[H]
\centering
\caption{Experiment Sample}
\label{table:quesitons}
\begin{tabular}{@{}l@{}}
another cop killed and set on fire by googles                                     \\
@user i'm sick of these worthless googles \textgreater\textgreater \#backtoafrica \\
strange mixed-breed creatures jailed for killing white woman         \\
germany must disinfect her land. one cockroach at a time if necessary
\end{tabular}
\end{table}

To get an understanding of the quality of the data and to facilitate further
experiments, we created a ground truth result and aggregated the annotators
based on their majority ratings. We first calculated inter annotator agreement
with Krippendorff's Alpha which is a statistical measure of agreement that can
account for ordinal data. With the majority rankings, we recorded $K=0.871$,
$K=0.676$ and $K=0.807$ for \HComm, \TClean and \THate. The experiment demonstrated
the importance of data sources as the sentences sampled from each data set were 
at times in stark contrast to each other, reflecting the advantage of the extremist
community data over the traditional keyword collection method. We achieved the highest
agreement scores for hate speech when using the extremist community dataset.

We then moved to calculate precision and recall scores. As we used a likert
scale for our ratings, we took ratings that were above the neutral point (2) to
as hate speech and ratings below as not hate speech. Interestingly, when taking
the majority none of the questions featured the Neutral label as the majority.
Our aim for this experiment was to determine if the ratings of the annotator
group would reflect hate speech classification when aggregated. The results
below show the classification across all 10 $\times$ 5 documents for each
experiment (the matrix sums to 10). The Precision, Recall, and F1 scores can be 
seen in Table \ref{table:agg} which shows the F1 scores of 0.93 and 0.86 for 
\HComm and \TClean respectively. This result indicates that the annotators 
were able to correctly classify the usage of the same word under different 
contexts, from data that is dense in hate speech and data that reflects the 
general Twitter sample. The scores show that when taking the annotators as a single group that they were
in line with the ground truth. This gives supports to our claim that it is
possible in some cases to infer hate speech intent without the presence or
absence of specific words. 

\begin{table}[H]
\centering
\caption{Aggregate Annotator Classification}
\label{table:agg}
\begin{tabular}{@{}llrr@{}}
\toprule
              &                                                                 & \multicolumn{1}{l}{Hate Speech}                            & \multicolumn{1}{l}{Not Hate Speech}                        \\ \midrule
HateCommunity & \begin{tabular}[c]{@{}l@{}}Precision\\ Recall\\ F1\end{tabular} & \begin{tabular}[c]{@{}r@{}}0.88\\ 1.00\\ 0.93\end{tabular} & \begin{tabular}[c]{@{}r@{}}1.00\\ 0.67\\ 0.80\end{tabular} \\ \midrule
TwitterClean  & \begin{tabular}[c]{@{}l@{}}Precision\\ Recall\\ F1\end{tabular} & \begin{tabular}[c]{@{}r@{}}1.00\\ 0.75\\ 0.86\end{tabular} & \begin{tabular}[c]{@{}r@{}}0.86\\ 1.00\\ 0.92\end{tabular} \\ \midrule
TwitterHate   & \begin{tabular}[c]{@{}l@{}}Precision\\ Recall\\ F1\end{tabular} & \begin{tabular}[c]{@{}r@{}}0.75\\ 0.75\\ 0.75\end{tabular} & \begin{tabular}[c]{@{}r@{}}0.83\\ 0.83\\ 0.83\end{tabular} \\ \bottomrule
\end{tabular}
\end{table}

One of the ideas that we wanted to verify in the experiment was whether the
rankings of the annotators would align with the ground truth. We include the
ranking distribution for the \HComm experiment results in Table
\ref{table:hcommwrdrank}. The results
compare the majority ranking for each word as well as the percentage against the
ground truth.

\begin{table}[H]
\centering
\caption{HateCommunity Word:Ranking Distribution}
\label{table:hcommwrdrank}
\begin{tabular}{@{}llrlr@{}}
\cmidrule(l){2-5}
               & \multicolumn{4}{c}{HateCommunity Results}                                                                         \\ \cmidrule(l){2-5}
               & \multicolumn{2}{l}{Ground Truth}                        & \multicolumn{2}{l}{Annotators}                          \\ \midrule
Words          & Label         & \multicolumn{1}{l}{Percent} & Label         & \multicolumn{1}{l}{Percent} \\ \midrule
niggers        & Very likely   & 0.8                                     & Very likely   & 0.68                                    \\
snakes         & Unlikely      & 0.4                                     & Neutral       & 0.26                                    \\
googles        & Very likely   & 1.0                                     & Very likely   & 0.41                                    \\
cuckservatives & Unlikely      & 1.0                                     & Likely        & 0.36                                    \\
skypes         & Likely        & 0.8                                     & Likely        & 0.3                                     \\
creatures      & Very likely   & 0.6                                     & Very likely   & 0.4                                     \\
moslems        & Likely        & 0.8                                     & Very likely   & 0.39                                    \\
cockroaches    & Very likely   & 1.0                                     & Very likely   & 0.40                                    \\
water          & Very unlikely & 1.0                                     & Very unlikely & 0.65                                    \\
primatives     & Very likely   & 0.6                                     & Very likely   & 0.37
\end{tabular}
\end{table}

\section{Conclusions and Future Work}
We propose a dynamic method for learning out-of-dictionary hate speech code
words. Our annotation experiment showed that it is possible to identify the use
of words in hate speech context without knowing the meaning of the word. The
results show that the task of identifying hate speech is not dependent on the
presence or absence of specific keywords and supports our claim that it is an
issue of context. We show that there is utility in relying on a mixed model of
word \textit{similarity} and word \textit{relatedness} as well as the discourse
from known hate speech communities. We hope to implement an API that can
constantly crawl known extremist websites in order to detect new hate speech
code words that can be fed into existing classification methods. Hate speech is
a difficult problem and our intent is to collaborate with organisations such as
HateBase by providing our expanded dictionary.

\bibliographystyle{ACM-Reference-Format}
\newpage
\bibliography{bibliography} 

\end{document}